\documentclass[10pt,twocolumn,letterpaper]{article}

\usepackage{cvpr}              

\usepackage{graphicx}
\usepackage{amsmath}
\usepackage{amssymb}
\usepackage{booktabs}
\usepackage{color} 
\usepackage{multirow} 
\usepackage{algorithm}  
\usepackage{algpseudocode}  
\usepackage{amsmath}  
\usepackage[pagebackref,breaklinks,colorlinks]{hyperref}

\usepackage[capitalize]{cleveref}
\crefname{section}{Sec.}{Secs.}
\Crefname{section}{Section}{Sections}
\Crefname{table}{Table}{Tables}
\crefname{table}{Tab.}{Tabs.}

\def\cvprPaperID{424} 
\def\confName{CVPR}
\def\confYear{2023}

\begin{document}

\title{Local-to-Global Panorama Inpainting for Locale-Aware Indoor Lighting Prediction}

\author{Jiayang Bai$^1$, Zhen He$^1$, Shan Yang$^1$, Jie Guo$^1$\footnotemark[1], Zhenyu Chen$^1$, Yan Zhang$^1$, YanwenGuo$^1$\\
$^{1}$Nanjing University, Nanjing, China\\
{\tt\small  \{jybai, hz,yangshan\}@smail.nju.edu.cn guojie@nju.edu.cn}\\
{\tt\small MF21330012@smail.nju.edu.cn \{zhangyannju, ywguo\}@nju.edu.cn}
}


\twocolumn[{
\renewcommand\twocolumn[1][]{#1}
\maketitle
\centering
\vspace{-0.4cm}
\includegraphics[width=\textwidth]{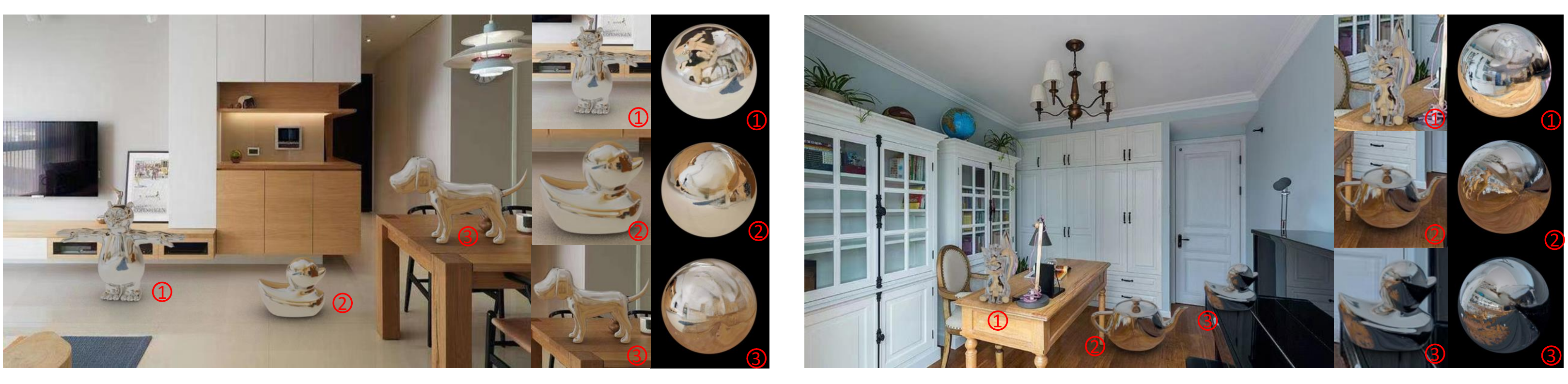}
\vspace{-0.5cm}
\captionsetup{type=figure}
\caption{We propose a locale-aware indoor illumination prediction method that can generate a full and texture-rich HDR panorama (the third column in each group) at any locale in the scene, enabling spatially-varying and consistent shading after virtual object insertion (the first and second columns).}
\label{fig:teaser}

\vspace{0.7cm}
}]
\footnotetext[1]{Corresponding author.}

\begin{abstract}
Predicting panoramic indoor lighting from a single perspective image is a fundamental but highly ill-posed problem in computer vision and graphics. To achieve locale-aware and robust prediction, this problem can be decomposed into three sub-tasks: depth-based image warping, panorama inpainting and high-dynamic-range (HDR) reconstruction, among which the success of panorama inpainting plays a key role. Recent methods mostly rely on convolutional neural networks (CNNs) to fill the missing contents in the warped panorama. However, they usually achieve suboptimal performance since the missing contents occupy a very large portion in the panoramic space while CNNs are plagued by limited receptive fields. The spatially-varying distortion in the spherical signals further increases the difficulty for conventional CNNs.
To address these issues, we propose a \emph{local-to-global strategy} for large-scale panorama inpainting. In our method, a depth-guided local inpainting is first applied on the warped panorama to fill small but dense holes. Then, a transformer-based network, dubbed \emph{PanoTransformer}, is designed to hallucinate reasonable global structures in the large holes. 
To avoid distortion, we further employ cubemap projection in our design of PanoTransformer. The high-quality panorama recovered at any locale helps us to capture spatially-varying indoor illumination with physically-plausible global structures and fine details. 
\end{abstract}


\section{Introduction}
Estimating high-dynamic range (HDR) indoor illumination from a single low-dynamic range (LDR), perspective image is a fundamental problem in computer vision and graphics. It has a huge impact on many applications including augmented and mixed reality (AR/MR) \cite{10.1111/cgf.12591}, scene understanding \cite{li2020inverse} and relighting \cite{https://doi.org/10.1111/cgf.14283}. 
However, this problem is highly ill-posed as the pixel intensities in an observed perspective image are a complex function of scene geometries, material properties and lighting distributions.

With the advent of deep learning and large-scale datasets, convolutional neural networks (CNNs) are now the de-facto architectures for solving this ill-posed problem. However, many existing CNNs only recover a single global environment map \cite{gardner2017learning,apple-hdr,zhan2020emlight}, or some other kinds of compact representations, \emph{e.g.}, Spherical Harmonics (SH) \cite{https://doi.org/10.1111/cgf.13561} and Spherical Gaussians (SG) \cite{10.1145/3338533.3366562,Zhan_2021_ICCV}, at the camera's viewpoint of the input perspective image. This completely ignores spatially-varying lighting effects that dominate indoor scenarios and hence results in unrealistic shading and shadows for inserted virtual objects. 

Several works have noticed this challenge and tried to recover the independent illumination at each locale (the 3D location of the inserted object). For instance, Song and Funkhouser \emph{et al.}~\cite{song2019neural} have suggested decomposing the indoor illumination estimation problem into three sub-tasks: depth-based image warping, LDR panorama inpainting and HDR panorama reconstruction. The success of this pipeline lies in the second sub-task which tries to recover missing contents for the incomplete panorama. However, directly applying existing convolution-based image inpainting networks \cite{iizuka2017globally,li2017generative,liu2019coherent,pathak2016context,yu2018generative} to the incomplete panorama is impractical due to (1) the hole regions of warped images are very large ($>60$\% in general), which may easy let many image inpainting methods fail; (2) panoramas have severe spatially-varying distortions that will mislead convolution-based feature extraction modules. 

More importantly, as linear operations have a limited receptive field, conventional convolutions fail to capture long-distance relationships that prevail in panoramas. To fill large holes in the warped panorama according to long-range dependencies, we resort to a transformer-based architecture, dubbed \emph{PanoTransformer}. 
Our PanoTransformer is specially designed for handling distorted spherical signals with cubemap projections and is adept at restoring visually plausible and globally consistent contents in the large out-of-view regions. 
We observe that many relatively small but dense holes are actually caused by pixel stretching during image warping at a given locale. Considering that these missing regions are visible in the input, we design a local inpainting module and apply it on the warped incomplete panorama before restoring global structures via PanoTransformer. Guided by estimated depth values, our local inpainting module can efficiently fill the gaps between sparse pixels without introducing any artifact. 
We also find that this local inpaining module helps our PanoTransformer better capture global structures in the panorama and make the training stable. Extensive experiments on multiple datasets verify that the proposed \emph{local-to-global panorama inpainting pipeline} significantly improves the quality of recovered indoor lighting distribution at any locale and enables high-fidelity and globally-coherent shading on inserted virtual objects. In particular, our method can reproduce fine texture details that are consistent with the inserting points on specular surfaces (see Fig. \ref{fig:teaser}).

In summary, we make the following contributions:
\begin{itemize}
    \item  A local-to-global panorama inpainting pipeline is proposed to fill missing contents in the warped panorama, enabling reproducing spatially-varying indoor illumination with physically-plausible global structures and fine details at any locale.
    \item A transformer-based network, named PanoTransformer, is designed to capture distortion-free global features in the panorama and restore globally-consistent structures accordingly. 
    \item A new large-scale panorama inpainting dataset is collected with paired masked input and ground truth panoramas.
\end{itemize}

\section{Related Work}

\textbf{Traditional methods for indoor lighting estimation. } 
To capture the lighting distribution at a target location in an indoor scene, a common practice is to use a physical probe~\cite{debevec2008rendering,debevec2012single,reinhard2010high} or other 3D objects~\cite{weber2018learning,georgoulis2017around,calian2018faces,yi2018faces}. However, inserting physical objects into the scene is not feasible in many scenarios. Moreover, it is usually expensive and has poor scalability. 
Another line of works tries to estimate illumination by jointly optimizing the geometry, reflectance properties, and lighting models of the scene. In order to find the set of values that best explained the observed input image, previous works~\cite{karsch2011rendering,karsch2014automatic,lombardi2015reflectance,zhang2016emptying} regressed the relevant parameters by minimizing the difference between the generated image and the ground truth. 

\textbf{Deep learning-based methods for indoor lighting estimation. } 
Recently, significant progress has been made in lighting estimation with deep learning. Some works~\cite{eigen2015predicting,gardner2017learning,lettry2018darn,eigen2014depth} directly trained convolutional neural networks to estimate scene parameters including albedos, normals, shadows and lights from a single image. EMLight~\cite{zhan2020emlight} inferred Gaussian maps to guide the illumination map generation. However, it simplified the scene
geometry to be a spherical surface and did not allow to predict spatially-varying lighting. Inspired by earth mover’s distance, GMLight~\cite{zhan2022gmlight} added a geometric mover’s loss to the pipeline in EMLight. 
Given a perspective image and a 2D location in that image, Garon \emph{et al.}~\cite{garon2019fast} achieved real-time spatially-varying lighting estimation by regressing a 5th order SH representation of the lighting at the selected location with CNNs. 
For estimating the incident illumination at any 3D location, Srinivasan \emph{et al.}~\cite{srinivasan2020lighthouse} predicted unobserved scene contents with a 3D CNN. Xu \emph{et al.} \cite{xu2020real} detected the planar surfaces in the scene and computed its overall illumination. Inspired by classic volume rendering techniques, Wang \emph{et al.}~\cite{wang2021learning} proposed a unified, learning-based inverse rendering framework that can recover 3D spatially-varying lighting from a single image. These works were plagued by the limited receptive fields and predicted inconsistent contents in those regions missing dued to pixel streching.

\begin{figure*}[t]
  \includegraphics[width=\textwidth]{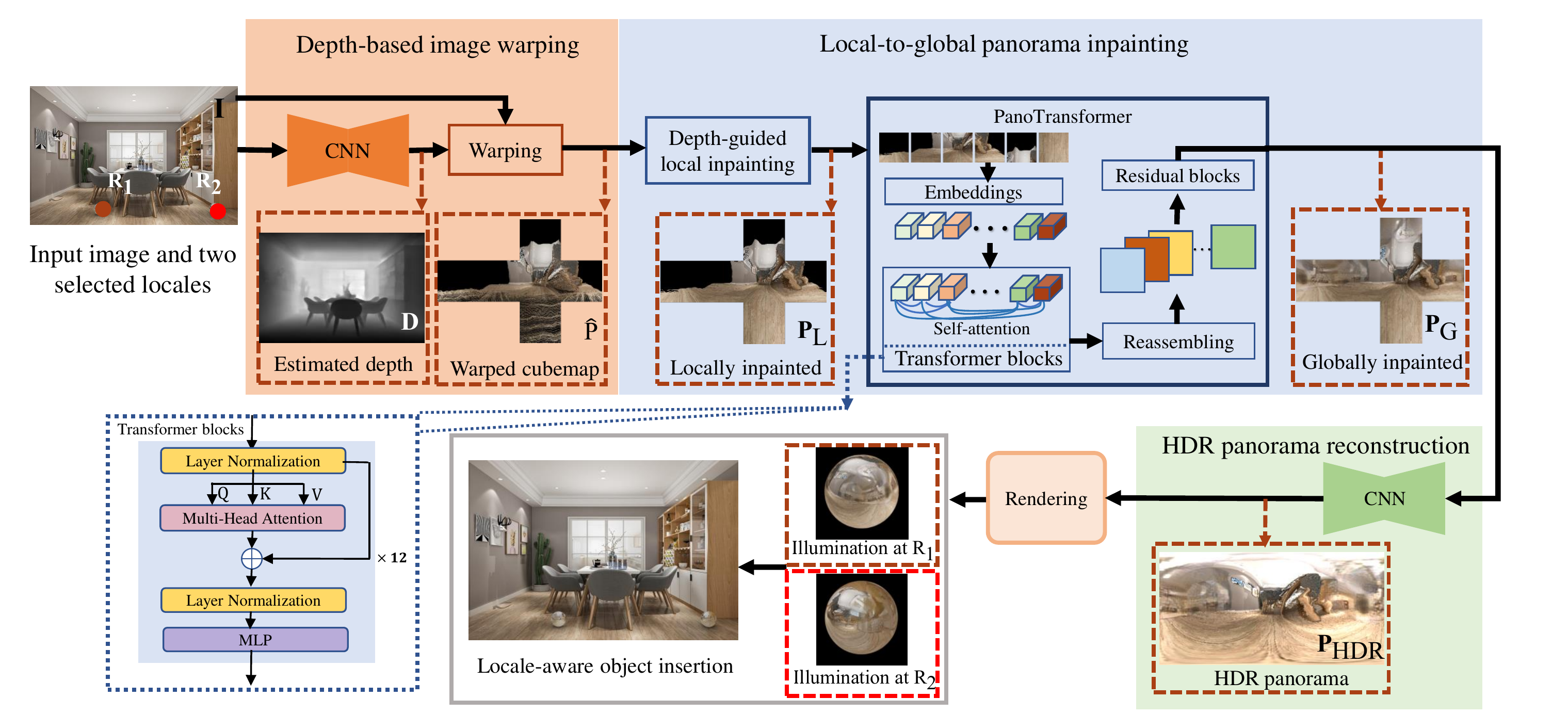}
  \caption{Architecture of the proposed locale-aware indoor illumination prediction method. The key is a local-to-global panorama inpainting pipeline that recovers both physically-plausible global structures and fine local details from a FOV-limited input image.}
  \label{fig:pipeline}
\end{figure*}
\section{Local-to-global Panorama Inpainting}
\label{section:overview}
Our goal in this paper is to estimate a full HDR environment map $\mathbf{P}_{\text{HDR}}$ at a  locale $\mathbf{R}$ of an FOV-limited LDR image $\mathbf{I}$.
We follow the general framework of Song \emph{et al.}~\cite{song2019neural} which decomposes this ill-posed problem into three sub-tasks: depth-based image warping, LDR panorama inpainting and HDR panorama reconstruction (see Fig. \ref{fig:pipeline}). In the first task, we leverage the recent DPT~\cite{dpt2021} to estimate the depth map $\mathbf{D}$ for $\mathbf{I}$. 
Then $\mathbf{D}$ and $\mathbf{I}$ are geometrically warped and transformed to $360^{\circ}$ spherical panoramas centered at the selected locale $\mathbf{R}$, denoting as $\mathbf{\hat{D}}$ and ${\hat{\mathbf{P}}}$ respectively. 
This warping operation is realized through a forward projection using the estimated scene geometry and camera pose. Some regions in $\mathbf{\hat{P}}$ are missing because they do not have a projected pixel in $\mathbf{I}$.
The following task, LDR panorama inpainting, aims to infer a dense panorama from the sparse ${\hat{\mathbf{P}}}$. In this section, we introduce a local-to-global panorama inpainting pipeline that can recover a complete LDR panorama $\mathbf{P}_{G}$ from $\mathbf{\hat{P}}$. 
In the last sub-task, we utilize the state-of-the-art network of ~\cite{LDR2HDR} to reconstruct the HDR panoramas $\mathbf{P}_{\text{HDR}}$ from $\mathbf{P}_{G}$ for inserted objects at $\mathbf{R}$. We detail our local-to-global pipeline in this paper and more details about the other off-the-shelf networks are provided in the supplemental materials.

\subsection{Observation and overview}

\begin{figure}[t]
  \begin{center}
  \renewcommand\tabcolsep{1.0pt}
  \begin{tabular}{ccc}
  \small{Input} &\small{Attention matrix} & \small{Attention map}  \\
    \includegraphics[height=0.24\linewidth, clip]{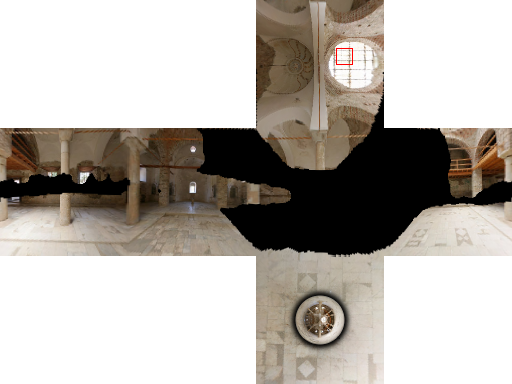}&
    \includegraphics[height=0.24\linewidth, clip]{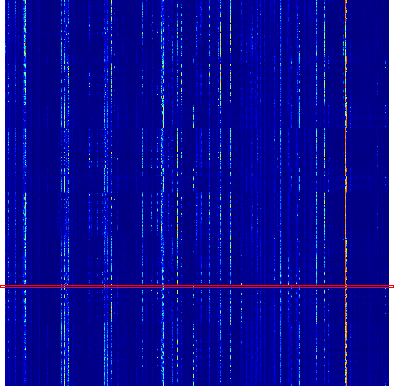}&
    \includegraphics[height=0.24\linewidth, clip]{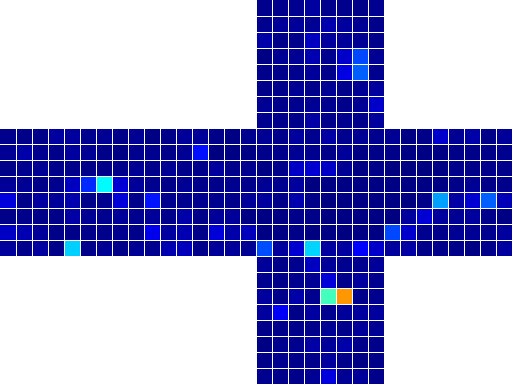} 
     \\
    
    \includegraphics[height=0.24\linewidth, clip]{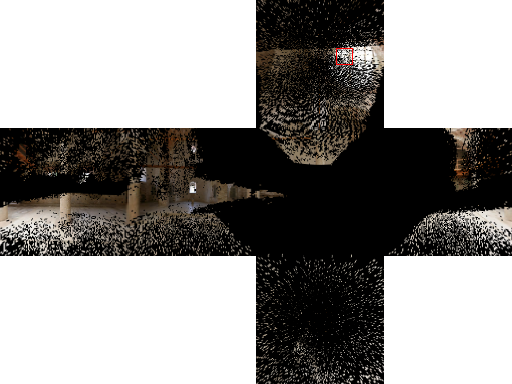}&
    \includegraphics[height=0.24\linewidth, clip]{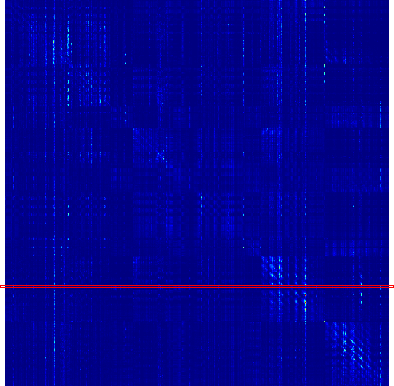}&
    \includegraphics[height=0.24\linewidth, clip]{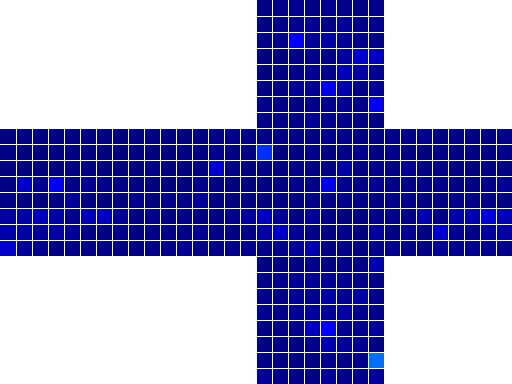}
  \end{tabular}
  \end{center}
  \caption{From left to right: the input incomplete panoramas with red boxes specifying the selected patches, attention matrices of the 9th transformer block with red lines indicating the attention scores of the selected patches, attention maps of the selected patches.}
  \label{fig:attention}
\end{figure}

For panorama inpainting, previous works \cite{song2019neural} mostly resort to fully convolutional networks. However, CNN-based models achieve suboptimal performance due to large-scale sparse missing contents in the warped panorama and some inherent limitations of convolutional layers. CNNs are good at preserving local structures and modeling rich textures but fail to complete the large hole regions. 
Therefore, previous works can hardly acquire sufficiently broad context as well as significant high-level representations from sparse omnidirectional images. Meanwhile, the distortion in spherical panoramas will further hamper the performance on large-scale inpainting.
    
Compared to CNN models with limited perspective fields, transformer is designed to support long-term interaction via the self-attention module~\cite{vit}. The self-attention mechanism can directly compute the interaction between any pair of patches, naturally capturing long-range dependencies and having a global receptive field at every stage. However, we observe that transformers work poorly on the sparse input which is the case in our task. We compare the attention maps and attention score maps for the selected patch from a sparse panorama and a dense one for illustration in Fig.~\ref{fig:attention}. As we can see, the query patch contains adequate illumination information. Given a dense input, the query patch impacts some regions (e.g. the ground) with red colors in the attention map. However, transformer blocks have trouble recovering the global structure from scattered pixels, thus the lighting can not pass information to invisible patches properly, resulting in the smooth attention map.

\begin{figure}[t]
  \includegraphics[width=\linewidth]{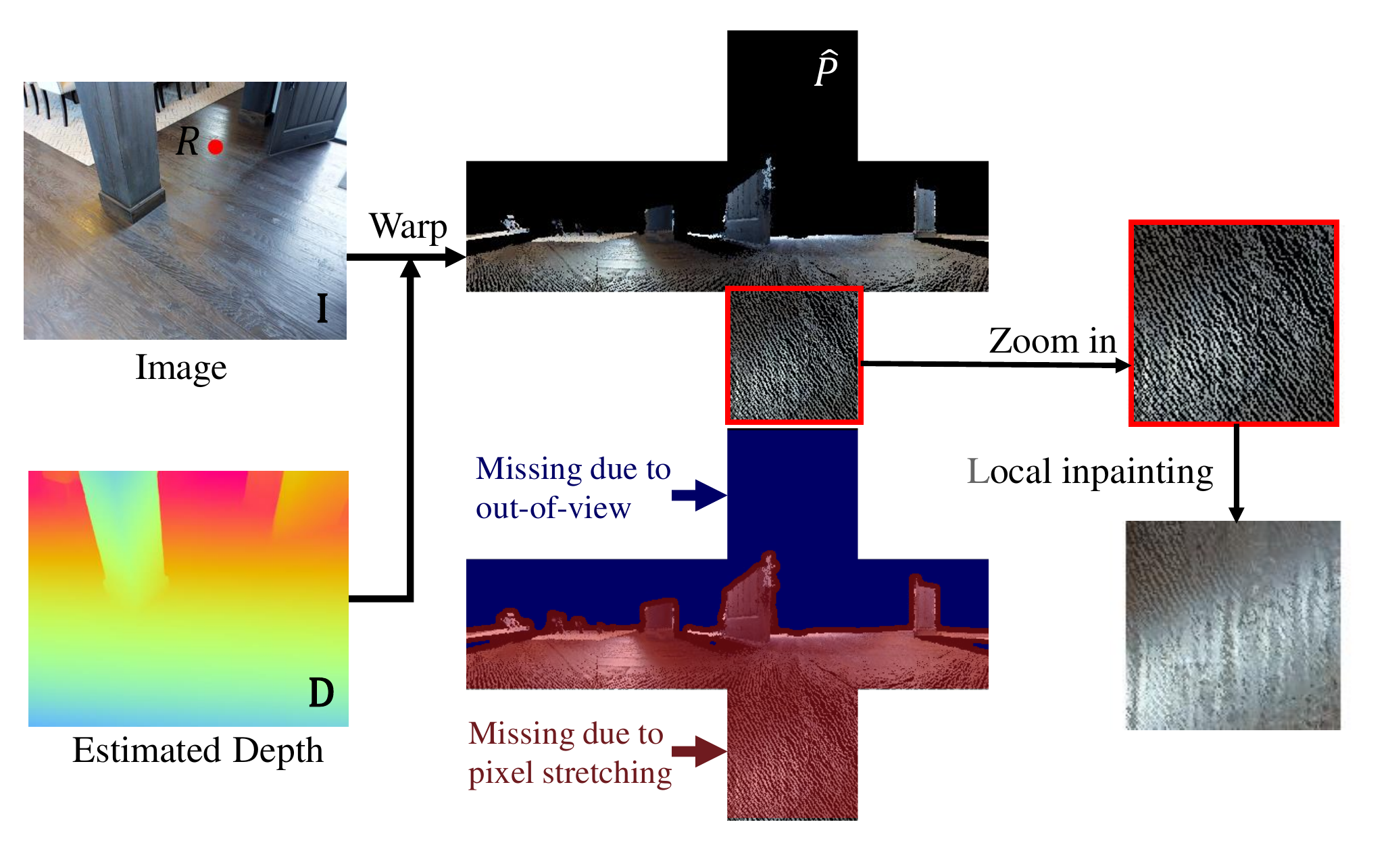}
  \caption{Illustration of warping. The warping operation results in missing regions which are categorized as pixel-stretching regions and out-of-view regions.}
  \label{fig:pixel}
\end{figure}

Unfortunately, the problem of sparsity is inevitable in our task due to the limited FOV of our input perspective images. The panorama captured at $\mathbf{R}$ should have a 360$^\circ$ FOV while some parts are out-of-view in $\mathbf{I}$. We mark the invisible regions with blue in Fig.~\ref{fig:pixel}. The missing parts are expected to be restored by a global understanding of the scene. Another factor of the sparsity stems from pixel stretching. During image warping, some regions marked in red in Fig.~\ref{fig:pixel} are actually visible in $\mathbf{I}$, but still have small holes due to pixel stretching. 

Based on the above observations, our main idea is that we first fill pixel-stretching regions according to their neighboring pixels to alleviate the sparsity, and then fill other large holes based on the global understanding of the whole scene. To this end, we propose a novel local-to-global inpainting pipeline, which can be formulated as follows:
\begin{equation}
    \mathbf{P}_{G} = \mathbf{M} \odot \mathbf{\hat{P}}+ (1-\mathbf{M})\odot \mathcal{G}(\mathcal{L}(\mathbf{\hat{P}});\mathcal{L}(\mathbf{M})) 
\label{equ:mask}
\end{equation}
where $\mathbf{M}$ is the binary mask indicating visible pixels in $\mathbf{\hat{P}}$ and $\odot$ denotes element-wise multiplication. 

The key to Eq.~\ref{equ:mask} is a local-to-global panorama inpainting pipeline that applies a local inpainting module $\mathcal{L}$ and a global inpainting module $\mathcal{G}$ sequentially on the warped panorama $\mathbf{\hat{P}}$.
Our local inpainting method aims to fill dense holes in the pixel-stretching regions, according to the depth information. The local inpainting module employs a modified bilateral filtering-based method to remove dense and small holes in the pixel-stretching regions. After that, a global inpainting module based on a novel transformer architecture is developed to extract reliable global features from visible regions and then fill large holes in the out-of-view regions. Our specially-designed transformer architecture, named PanoTransformer, takes a cubemap projection as input to resolve the problem of spatial distortion in the spherical signals.

\subsection{Depth-guided local inpainting}
\begin{algorithm}[tb]
  \caption{Depth-guided local inpainting}
  \label{code:localinpaint}
  \begin{algorithmic}[1]
    \Require
      The input image, $\mathbf{I}$;
      The estimated depth, $\mathbf{D}$;
      The warped image centered at selected location, $\mathbf{\hat{P}}$;
      The warped depth centered at selected location, $\mathbf{\hat{D}}$;
      The threshold $t$
    \Ensure
      The panorama inpainted locally, $\mathbf{P}_{L}$;
    \State $\mathbf{P}_{L} = \mathbf{\hat{P}}$
    \State $//$ Depth recovery.
    \State $\mathbf{\hat{D}} = \text{morphologyClosing}(\mathbf{\hat{D}})$
    \State $\mathbf{\hat{D}} = \text{bilateralFilter}(\mathbf{\hat{D}})$
    
    \State $//$ Depth-guided inpainting.
    \For{each $d\in \mathbf{\hat{D}}$}
      \State Calculate the pixel coordinate $\mathbf{c}_{p}$ of $d$ ;
      \State $\mathbf{c}_w = \text{PixelToWorldCoordinate}(\mathbf{c}_{p})$
      \State Project $\mathbf{c}_w$ to $\mathbf{D}$ to get its pixel coordinate $\mathbf{c}$;
      \State $d_i = \mathbf{D}[\mathbf{c} ]$
      \If {$\|d_i - d \| < t$}
        \State $\mathbf{P}_{L}[\mathbf{c}_p]=\mathbf{I}[\mathbf{c} ]$
      \EndIf
    \EndFor
  \end{algorithmic}
\end{algorithm}

The local inpainting module aims to alleviate the sparsity caused by pixel stretching. Our local inpainting method contains two basic steps shown in Algorithm~\ref{code:localinpaint}. First, we leverage morphology operations (Line 3) and bilateral filtering (Line 4) to fill holes in the warped depth $\mathbf{\hat{D}}$ as much as possible. The intuition of performing depth recovery is that depth values in the pixel stretching areas vary smoothly, while the warped panorama $\mathbf{\hat{P}}$ may have rich textures. 

Then, we traverse all valid depth values in $\mathbf{\hat{D}}$, and fill missing regions in $\mathbf{\hat{P}}$ with re-projected pixel values from $\mathbf{I}$. Guided by the recovered depth, we only fill those pixel-stretching regions with colors in $\mathbf{I}$. Note that our local inpainting method effectively fills dense but small holes and maintains the local structures.

\subsection{Global inpainting with PanoTransformer}
\label{section:lt}
For global inpainting, we design and train a transformer-based network, named PanoTransformer, to hallucinate contents in out-of-view regions. PanoTransformer can be logically separated to an encoder and a decoder, where the encoder captures long-range distortion-free representations, while the decoder gradually recovers the spatial information for generating accurate pixel-level predictions.

Considering the distortion in the equirectangular projection, PanoTransformer takes the warped cubemap projection as input. To distinguish pixels needed to be regressed, a binary mask is concatenated to the RGB channels in cubemap projection. 
We denote the input as $\mathbf{x}\in \mathbb{R}^{6\times H\times W\times 4}$ where $H$ and $W$ is the height and width of each face in the cubemap projection.

In the encoder, the input cubemap projection $\mathbf{x}$ is first reshaped into a sequence of flattened 2D patches $\mathbf{x}_p\in \mathbb{R}^{N \times(p^2\times 4)}$, where $(p, p)$ is the resolution of each image patch, and $N = \frac{6\times H\times W}{p^2}$ is the resulting number of patches. 
Each patch is then mapped to 1D tokens $z_i\in \mathbb{R}^d$ using a trainable linear projection and augmented with position embeddings to retain positional information.
These tokens are then fed into several transformer blocks. Each transformer block comprises multiheaded self-attention (MHSA)~\cite{NIPS2017_3f5ee243}, layernorm (LN)~\cite{ba2016layer} and MLP blocks.
In $l$-th transformer block, $\boldsymbol{z}^l$ is fed as the input, yielding $\boldsymbol{z}^{l+1}$ as follows:
\begin{equation}
    \boldsymbol{w}^l =\text{MHSA}(\text{LN}(\boldsymbol{z}^l)) +\boldsymbol{z}^l
\end{equation}
\begin{equation}
    \boldsymbol{z}^{l+1} =\text{MLP}(\text{LN}(\boldsymbol{w}^l)) +\boldsymbol{w}^l
\end{equation}

Noted that the token-dimensionality is fixed throughout all layers. Thus the resulting $\boldsymbol{z}$ can be transformed back to a cubemap-shape representation $\boldsymbol{z}\in \mathbb{R}^{6\times H\times W\times C}$ according to the position of the initial patch. $C$ is the channel of the extracted feature. 
These reconstructed priors $\boldsymbol{z}$ contain ample cues of global structure and coarse textures, thanks to the transformer’s strong representation ability and global receptive fields. 
$\boldsymbol{z}$ can be regarded as six images of the cubemap projection. We feed these image-like features to six Residual blocks~\cite{resnet} to replenish texture details. 
\begin{figure}[tb]
  \includegraphics[width=1.0\linewidth]{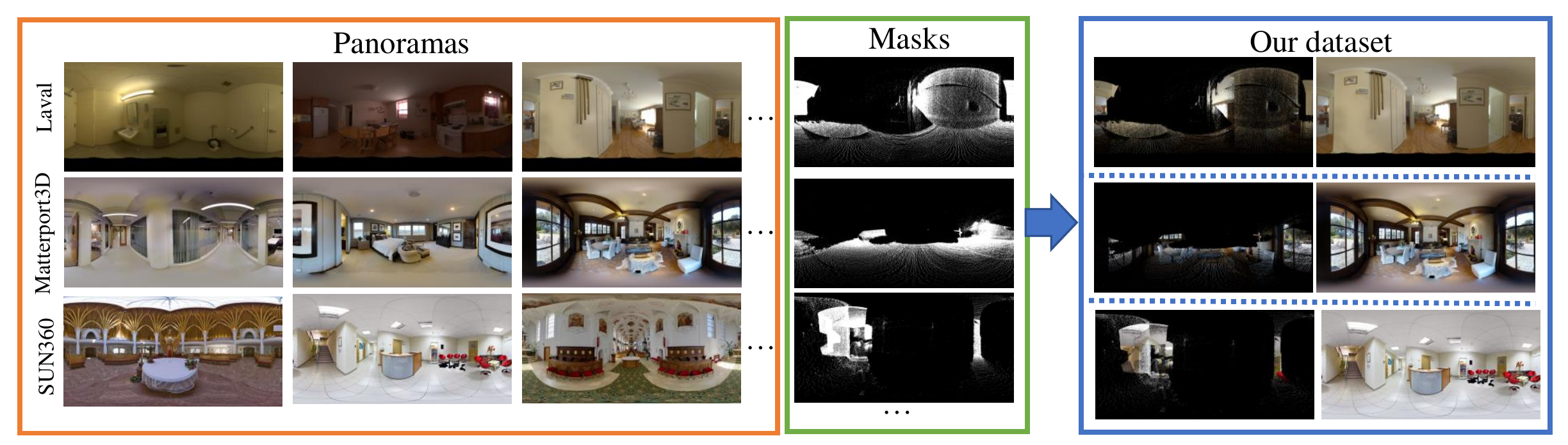}
  \caption{We apply masks from ~\cite{song2019neural} on high-quality panoramas from Matterport3D~\cite{Matterport3D}, SUN360~\cite{DBLP:conf/cvpr/XiaoEOT12} and Laval~\cite{gardner2017learning} to generate pairs of masked input and the ground truth for training.}
  \label{fig:dataset}
\end{figure}

\begin{figure}[tb]
\begin{center}
  \renewcommand\tabcolsep{1.0pt}
  \begin{tabular}{ccccccc}
  \multicolumn{4}{c}{\includegraphics[width=0.52\linewidth, clip]{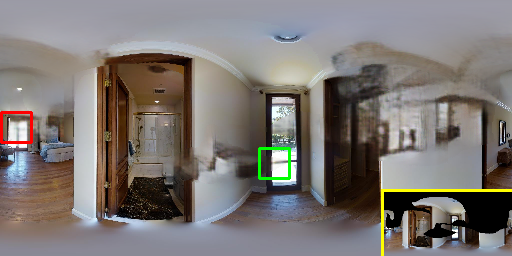}}&\includegraphics[width=0.13\linewidth,  clip]{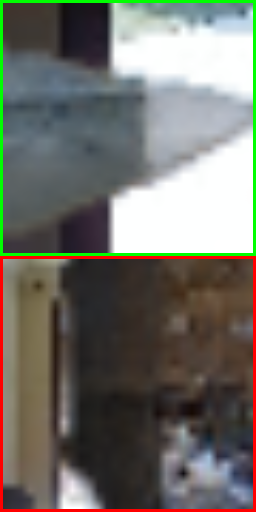}&
  \includegraphics[width=0.13\linewidth,  clip]{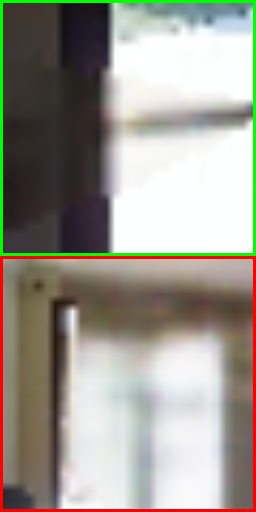}&
  \includegraphics[width=0.13\linewidth,  clip]{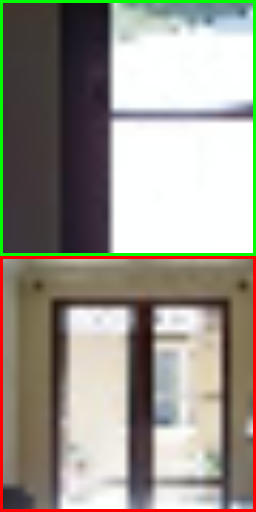}\\
  \multicolumn{4}{c}{\small{Our result and the input}}&\small{With~\cite{song2019neural}}&
  \small{Ours}&
  \small{Reference}\\
  \end{tabular}
  \end{center}
  \caption{Comparison between models trained with our dataset and the dataset of Song \emph{et al.}~\cite{song2019neural}.}
  \label{fig:nodata}
\end{figure}

\subsection{Dataset}
\label{section:dataset}
Currently, the only dataset that contains paired LDR perspective images and the corresponding HDR panoramas for a diverse set of locales is proposed by Song \emph{et al.}~\cite{song2019neural} based on Matterport3D~\cite{Matterport3D}. Unfortunately, the reconstructed HDR panoramas have obvious artifacts (\emph{e.g.}, stitching seams and broken structures) as we explained in the supplemental materials. This prohibits our model from inferring complete and globally consistent structures seen at each locale.

Considering the above issue, we collect a large scale dataset with high-quality and diverse panoramas from Matterport3D~\cite{Matterport3D}, SUN360~\cite{DBLP:conf/cvpr/XiaoEOT12} and Laval~\cite{gardner2017learning}.
Apart from the panoramas, training PanoTransformer also requires masks to generate the sparse input $\mathbf{\hat{P}}$. As the invisible regions are mainly on the top of panorama, we generate masks from the dataset of Song \emph{et al.}~\cite{song2019neural} instead of generating randomly. These sparse masks are obtained by geometrically warping, fitting to the real-world data distribution. These masks are locally inpainted before feeding to PanoTransformer.
The main difference of our dataset to Song \emph{et al.}'s~\cite{song2019neural} dataset is that our panoramas and masks are unpaired, hence we can randomly apply diversified irregular masks on one panorama to generate various input. Since we focus on the inpainting task, we do not require that the mask and the panorama are physically correlated. Our dataset (some examples are shown in Fig. \ref{fig:dataset}) ensures that the ground-truth panoramas are free from artifacts. In all, we gather 38,929 high-quality panoramas accompanied by randomly selected masks for training and 5,368 for evaluation. We ensure that the scenes used in the evaluation would not appear in the training procedure. 
As shown in Fig. \ref{fig:nodata}, the model trained with our dataset produces much better results than that trained with Song \emph{et al.}'s~\cite{song2019neural} dataset. Note the cluttered structures generated by the latter model due to artifacts in Song \emph{et al.}'s~\cite{song2019neural} dataset. Please refer to the supplemental material for more comparisons.

\subsection{Loss function and training details}
We optimize PanoTransformer by minimizing a pixel-wise reverse Huber loss ~\cite{Laina_2016_3DV} between the predicted panoramas and corresponding ground truth. Since using a standard L1 loss function to learn a binary light mask heavily penalizes even small shifts of a light source position, the reverse Huber loss takes advantage of L1 loss and L2 loss as below:
\begin{equation}
L_B=\left\{
\begin{aligned}
 &|y-\hat{y}|, & |y-\hat{y}| \leq T \\
&\frac{(y-\hat{y})^2+T^2}{2T},  & |y-\hat{y}| \ge T
\end{aligned}
\right.
\label{eq6}
\end{equation}
where $y$ is the ground truth value and $\hat{y}$ is the prediction. The threshold $T$ is set to 0.2 in our experiments.
To generate more realistic details, an extra adversarial loss is also involved in the training process. Our discriminator adopts the same architecture as Patch-GAN~\cite{patchgan}.

We implement our PanoTransformer using the PyTorch framework~\cite{pytorch}. Adam~\cite{kingma2017adam} optimizer is used with the default parameters $\beta_1=0.9$ and $\beta_2 = 0.999$ and an initial learning rate of 0.0001.
PanoTransformer is trained on our dataset for 100 epochs. Training are conducted on two NVIDIA RTX 3090 GPUs with a batch size of 8.

\begin{figure*}
  \includegraphics[width=\textwidth]{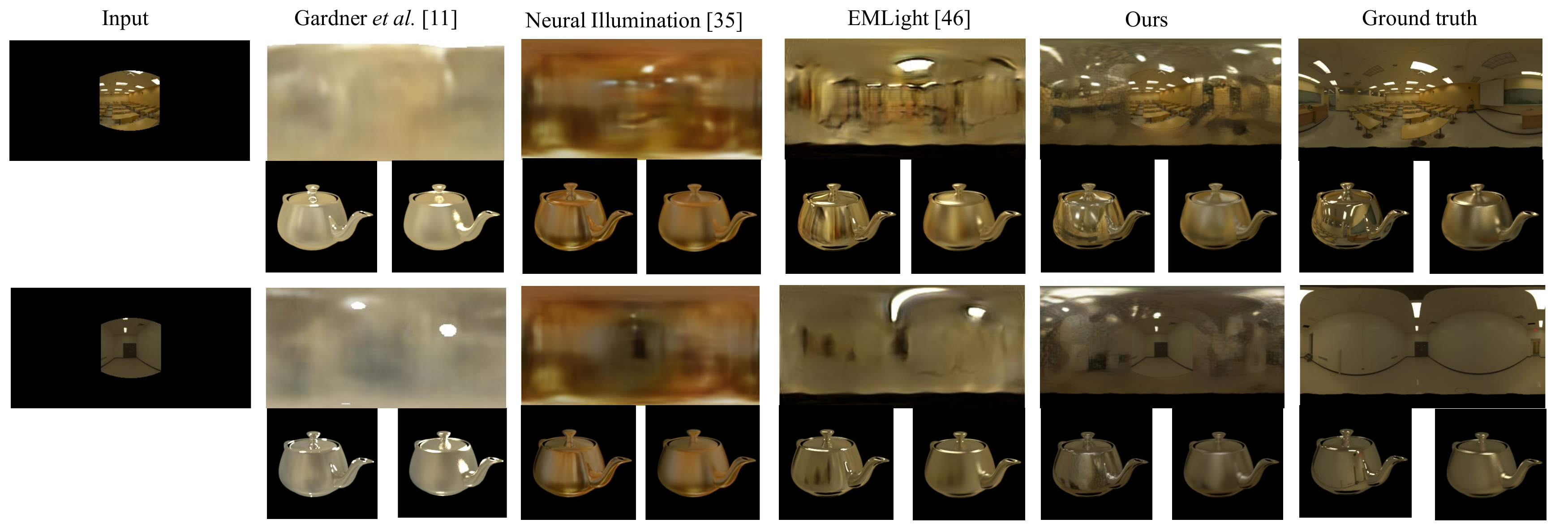}
  \caption{Rendering comparison with Gardner \emph{et al.}~\cite{Gardner17}, Neural Illumination~\cite{song2019neural} and EMLight~\cite{zhan2020emlight}. We render two teapots with a matte silver and a mirror material below each recovered/ground-truth illumination map. }
  \label{fig:ball_comparions}
\end{figure*}

\section{Experiments}
In this section, we evaluate the performance of our method in indoor illumination prediction and make comparisons with the state of the art. To further validate our local-to-global inpainting method, we also compare the inpainted results $\mathbf{P}_{G}$ against results from Neural Illumination~\cite{song2019neural}. More results and comparisons can be found in the supplemental materials.

\subsection{Evaluation for illumination estimation}

\begin{figure}[tb]
  \begin{center}
  \renewcommand\tabcolsep{1.0pt}
  \begin{tabular}{ccc}
    \includegraphics[width=0.32\linewidth, clip]{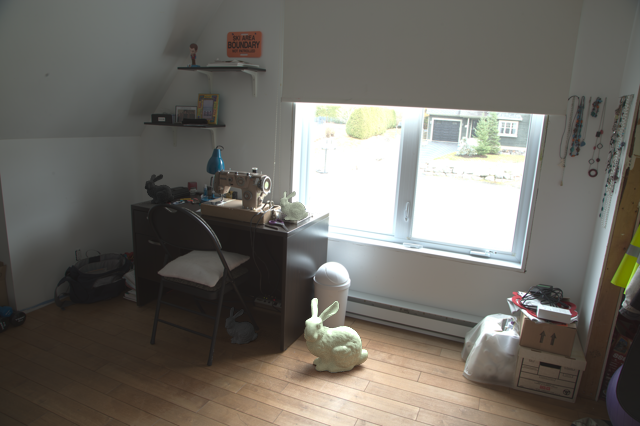}   &
    \includegraphics[width=0.32\linewidth, clip]{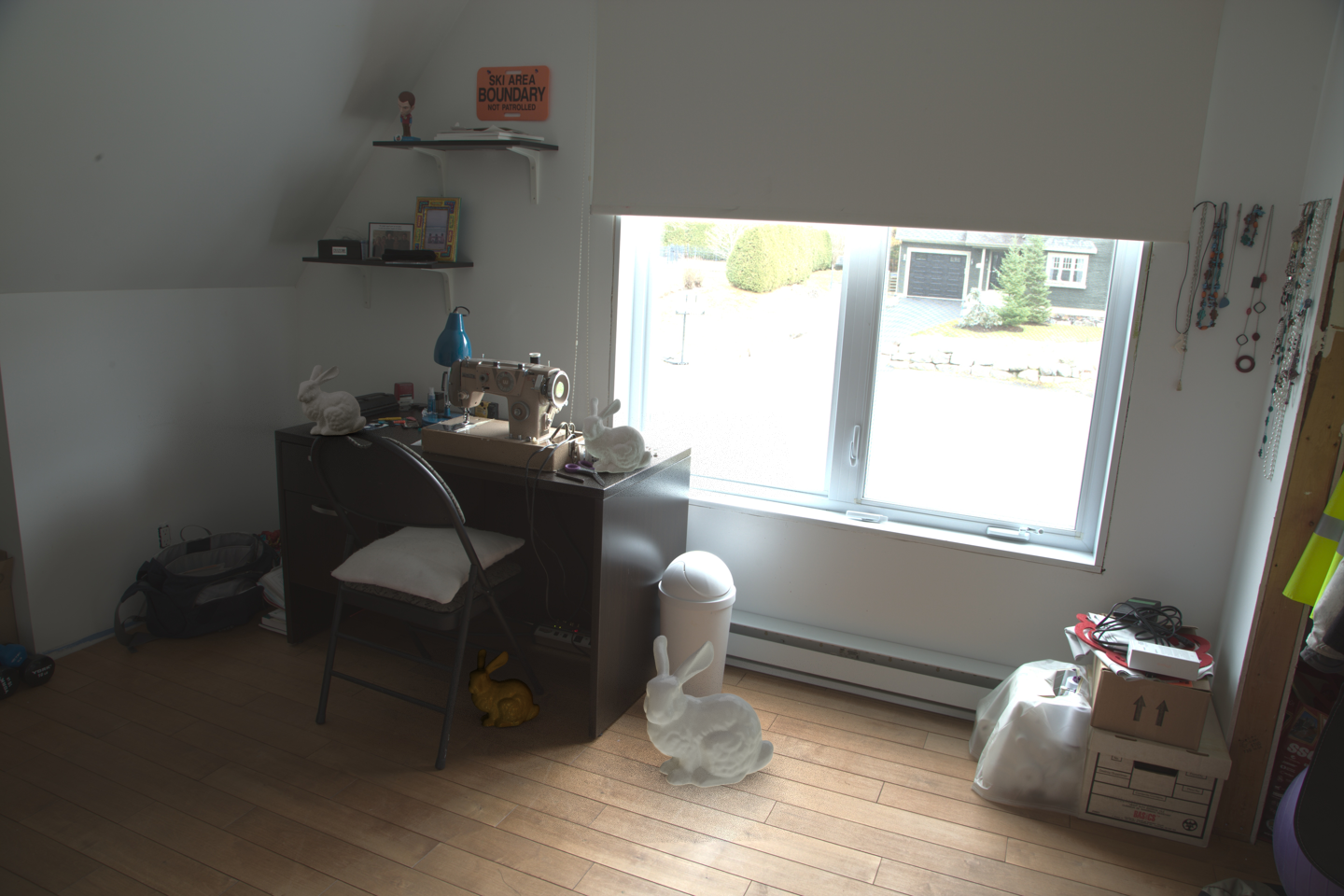} &
    \includegraphics[width=0.32\linewidth, clip]{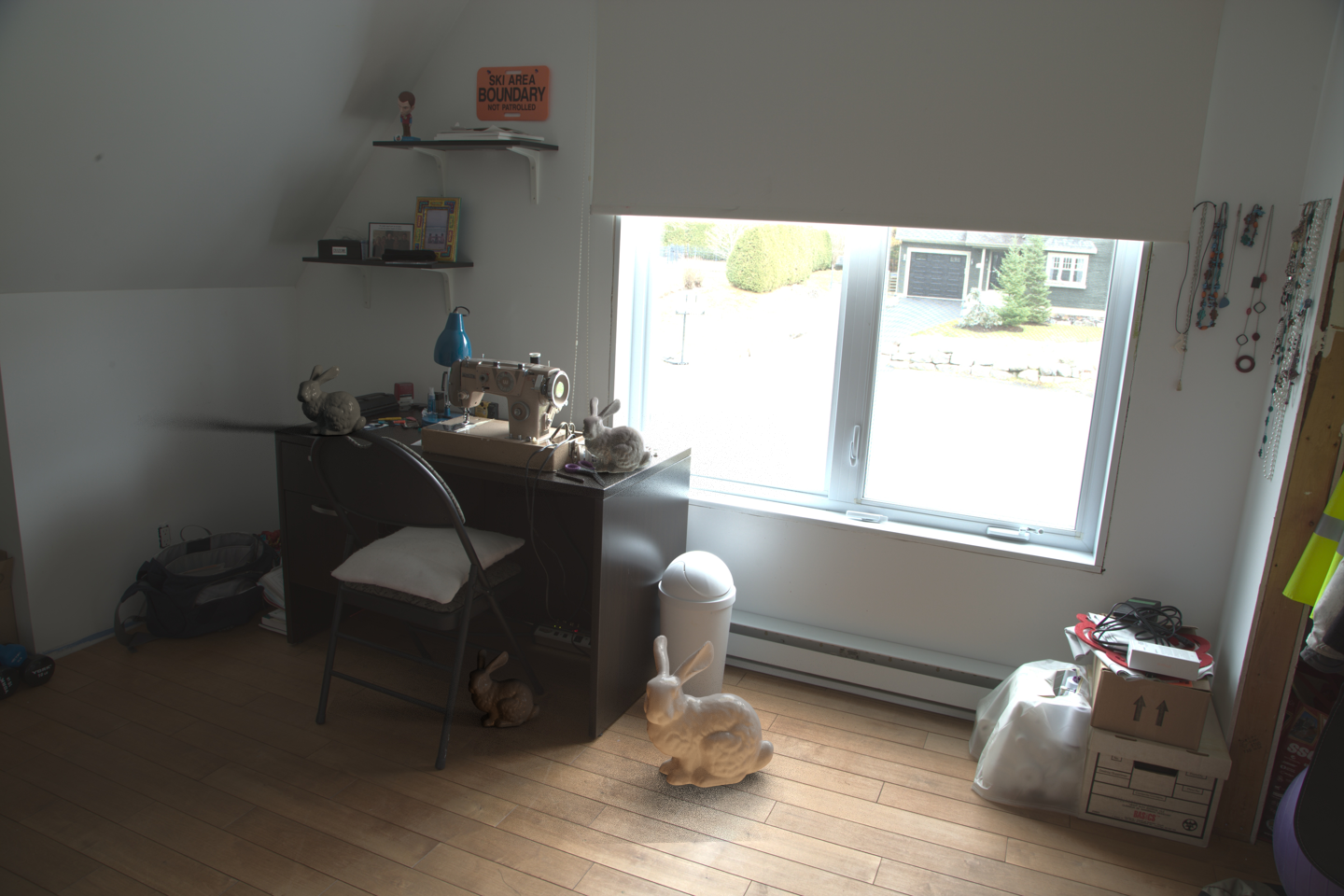}\\
    \includegraphics[width=0.32\linewidth,   clip]{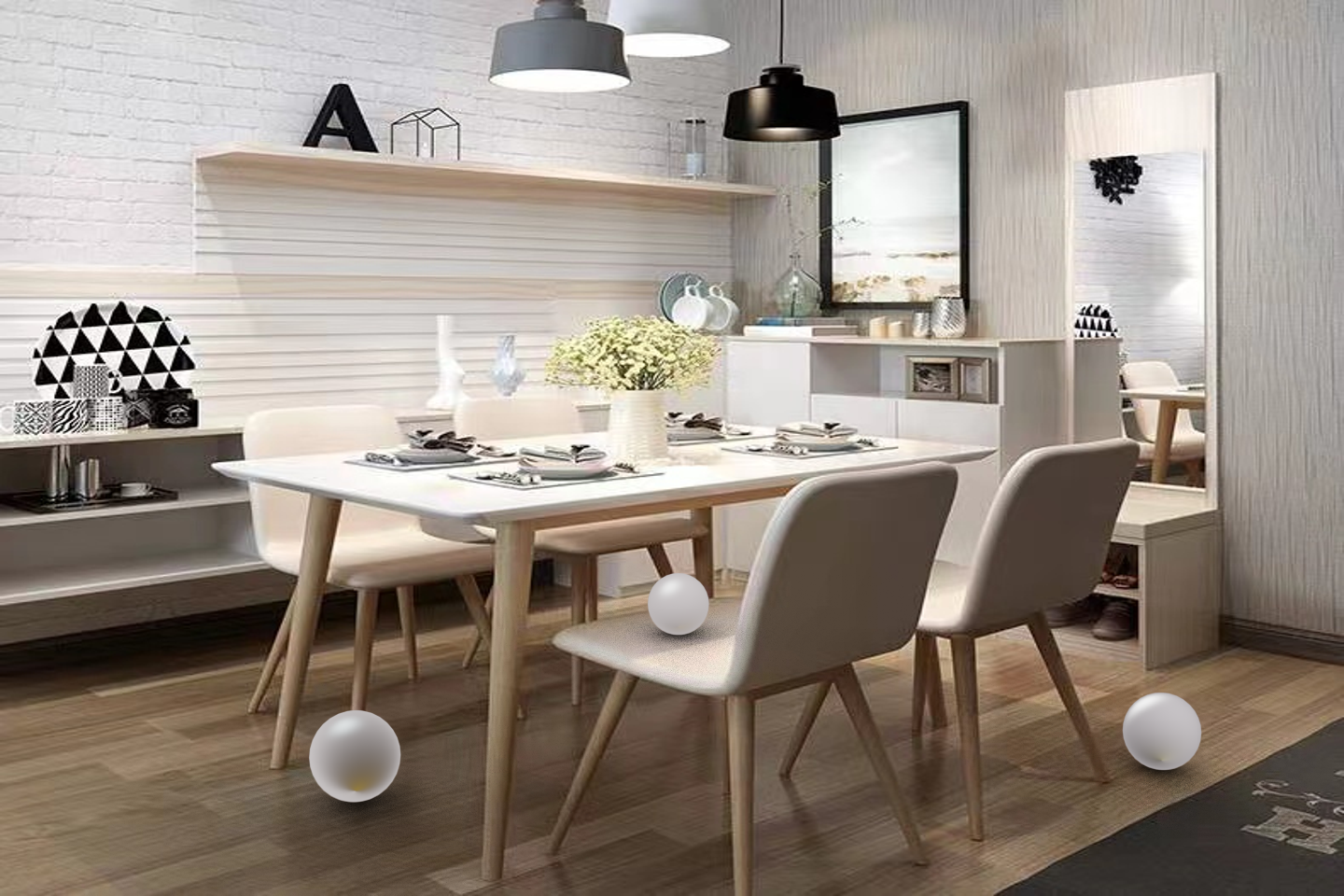}   &
    \includegraphics[width=0.32\linewidth,   clip]{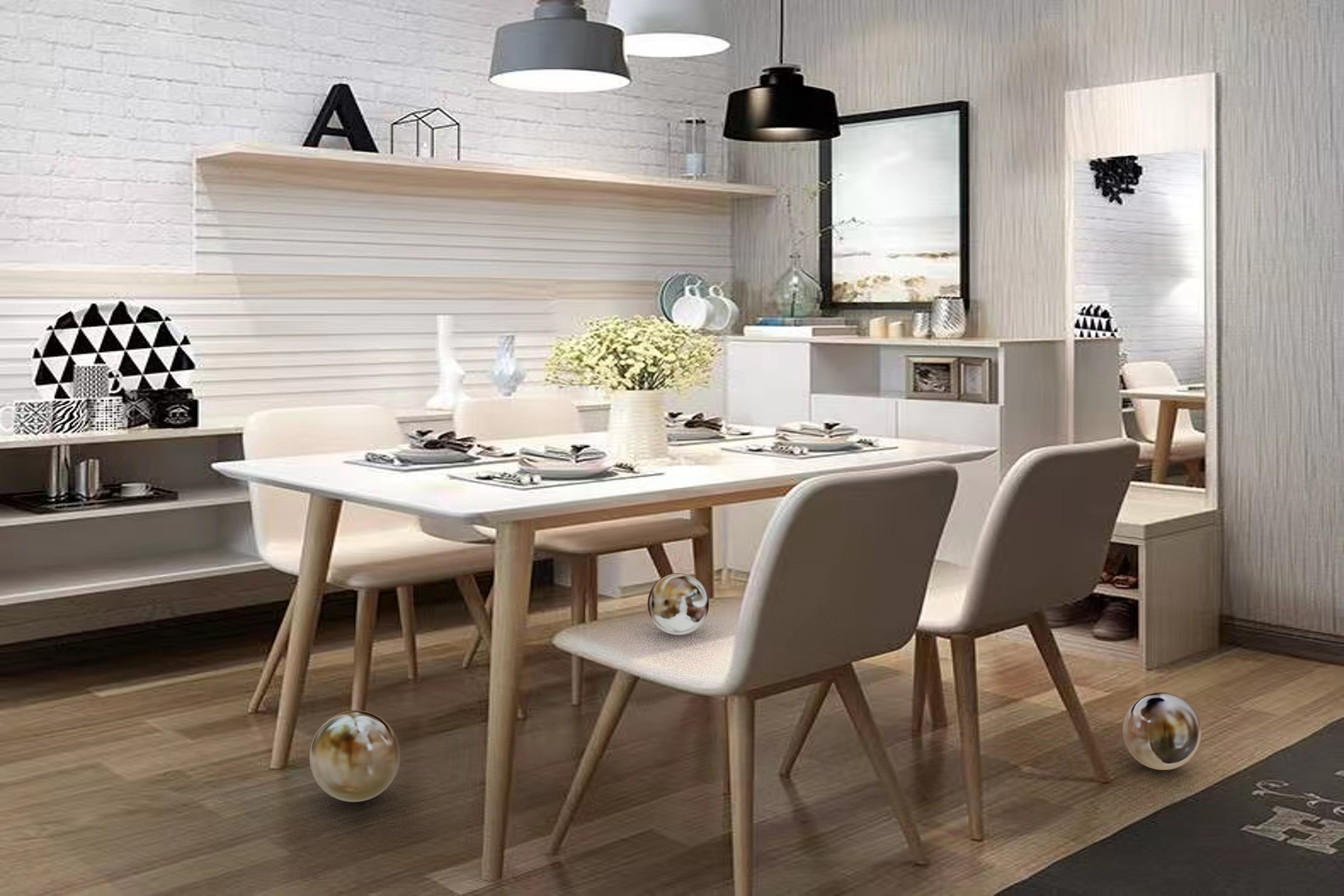} &
    \includegraphics[width=0.32\linewidth,   clip]{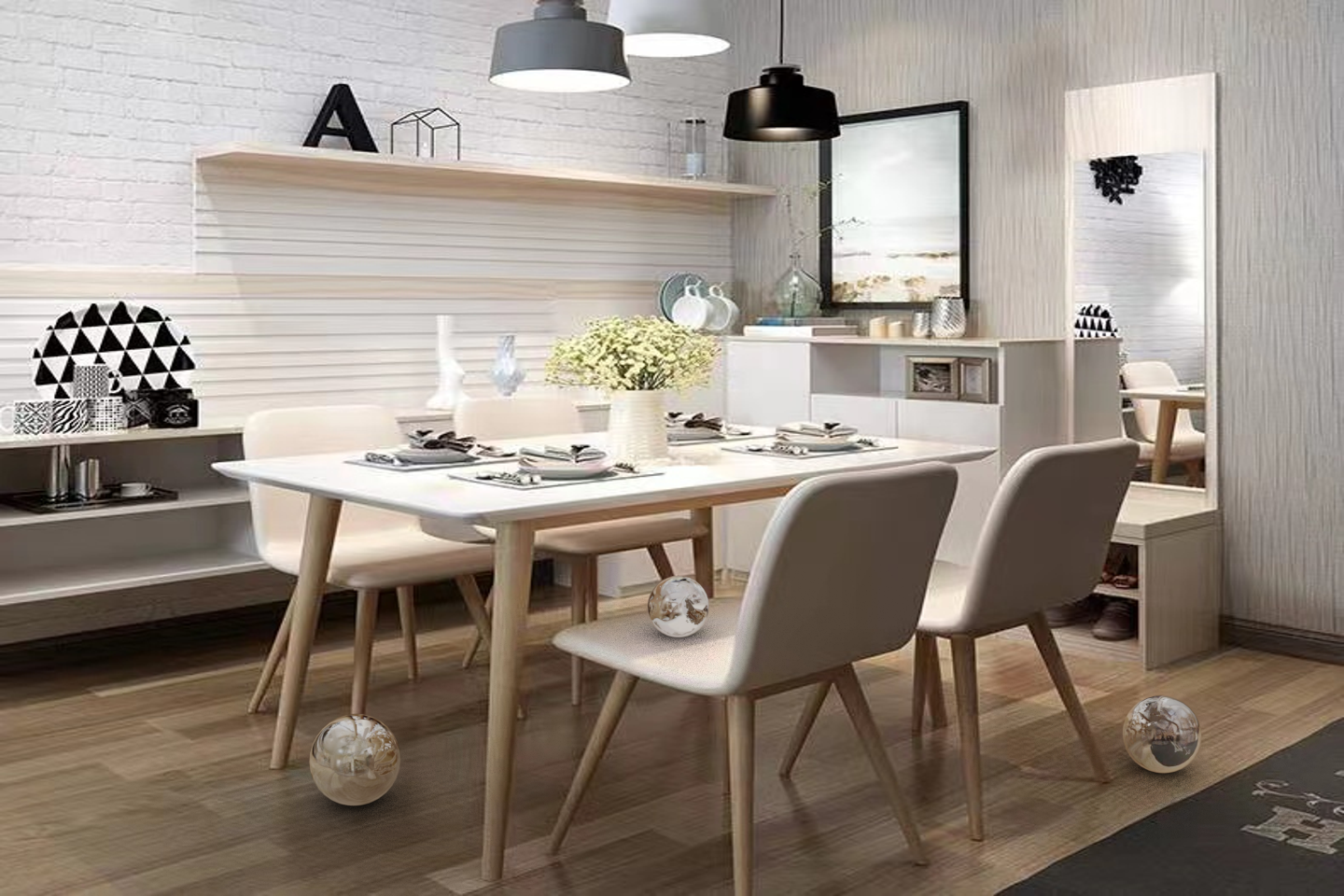} \\
    \includegraphics[width=0.32\linewidth, clip]{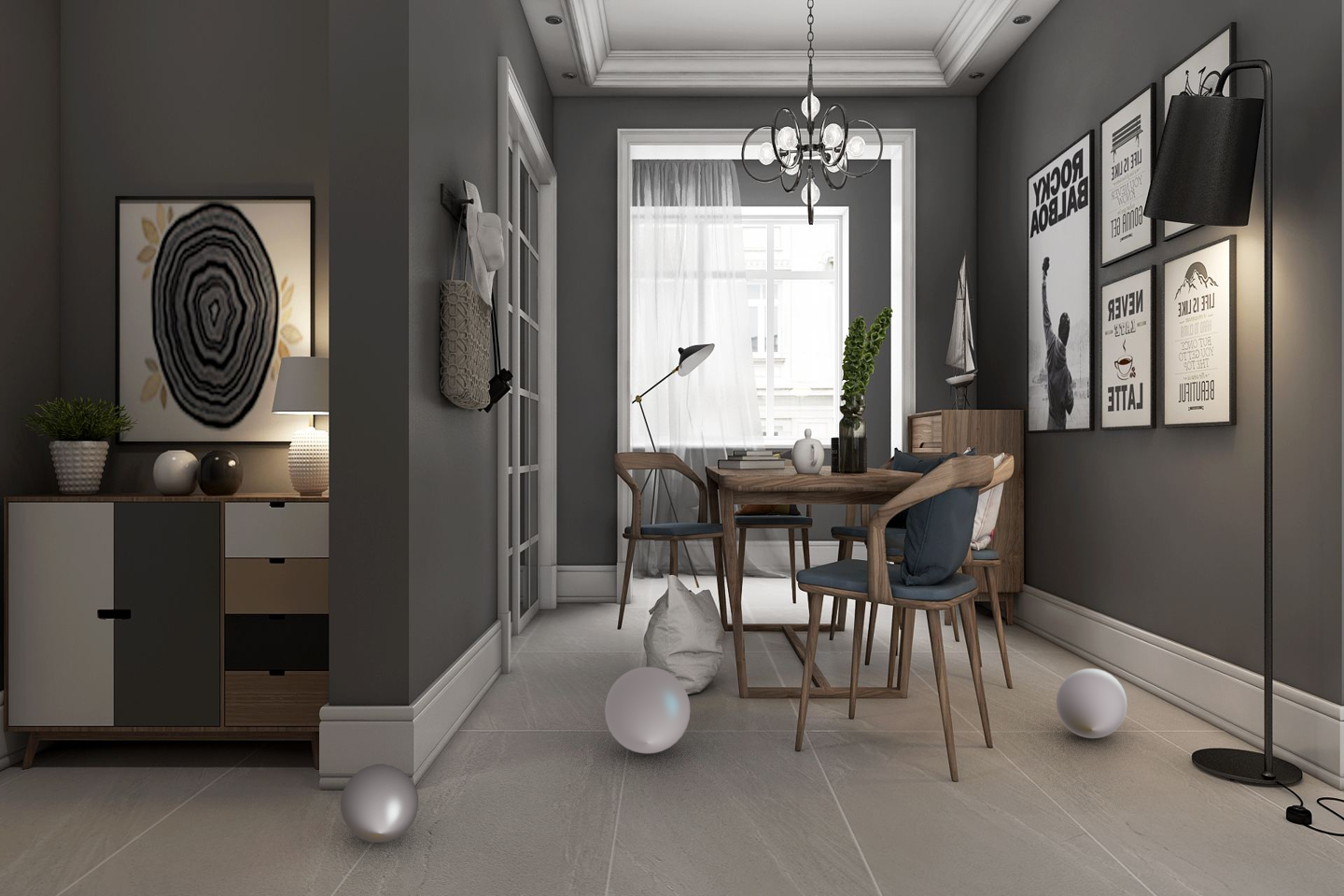}   &
    \includegraphics[width=0.32\linewidth, clip]{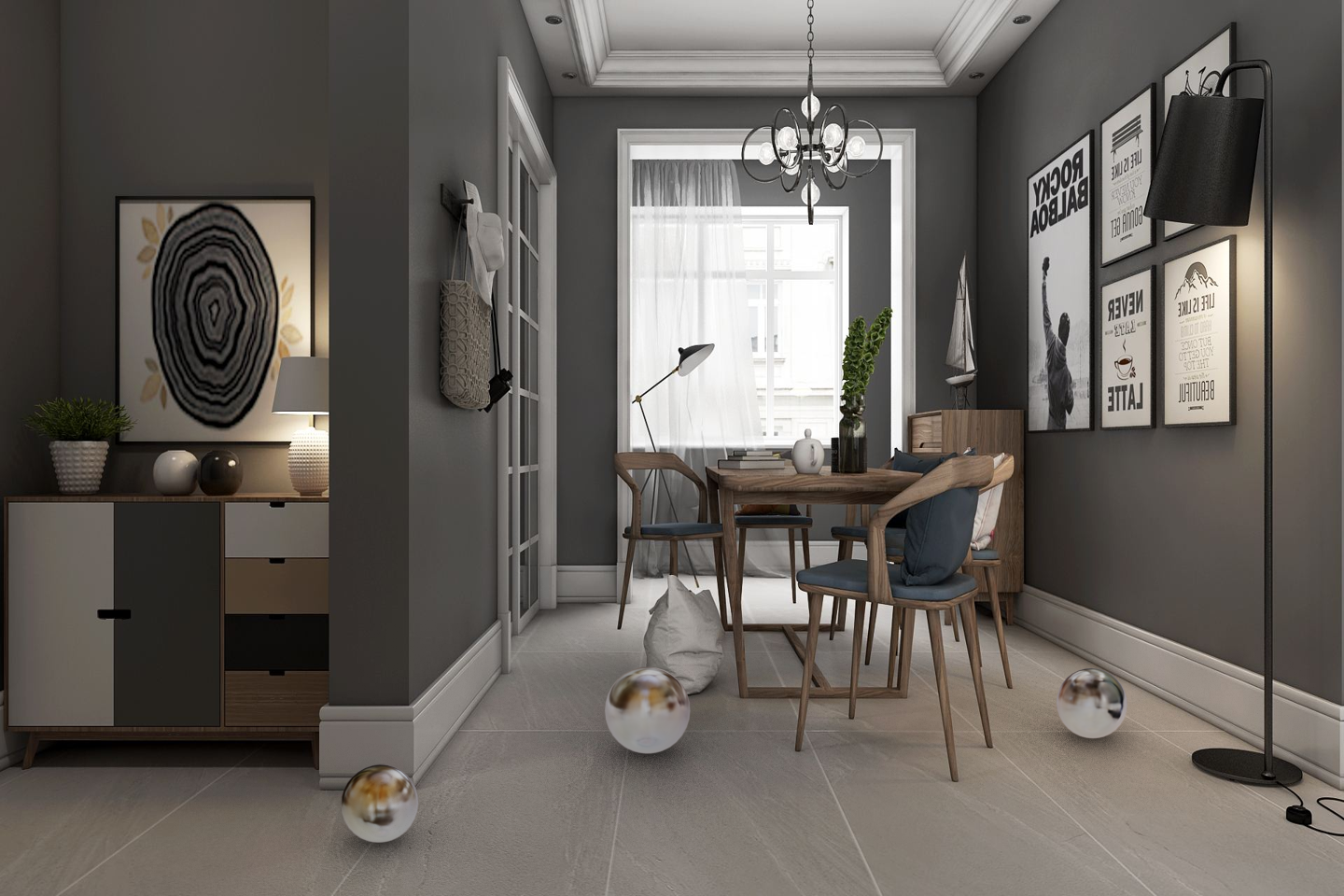} &
    \includegraphics[width=0.32\linewidth,   clip]{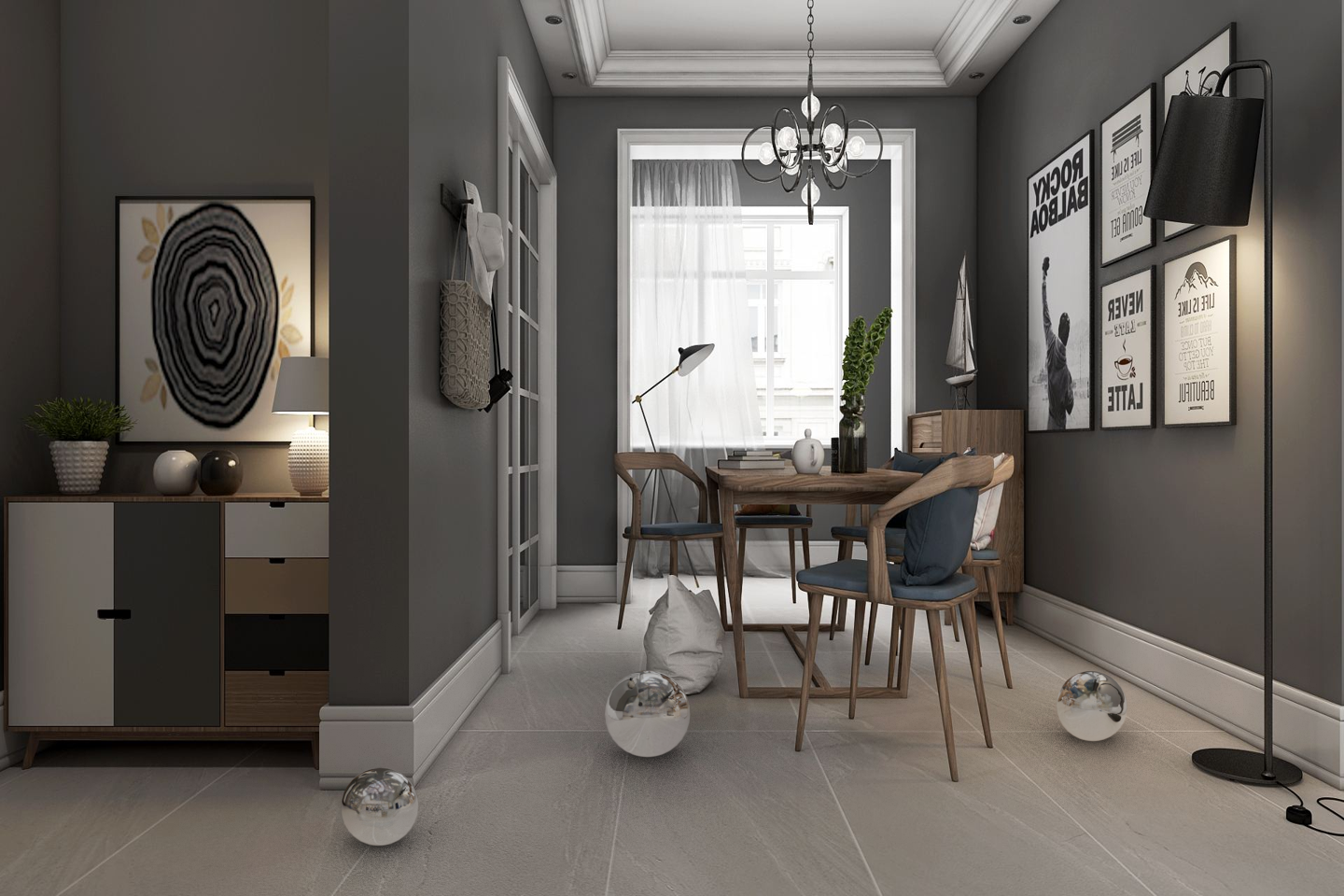} \\
    \small{Li \emph{et al.}~\cite{li2020inverse}}&\small{~\cite{song2019neural}}&\small{Ours}
  \end{tabular}
  \end{center}
  
  \caption{Results of inserted objects of different materials with estimated illumination by different methods. }
  \label{fig:inserted}
\end{figure}
\begin{figure}[tb]
  \includegraphics[width=\linewidth]{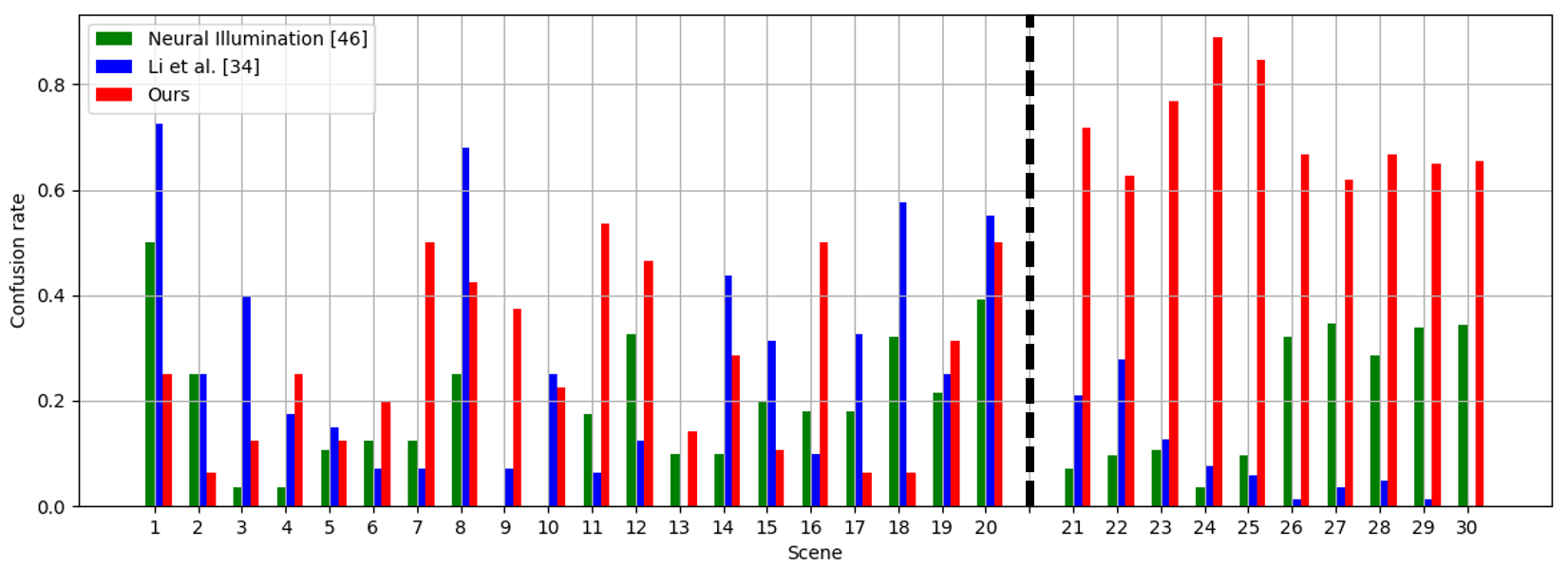}
  \caption{Results of user study on 30 scenes. Each scene in the Garon \emph{et al.}'s dataset~\cite{garon2019fast} (Scenes 1-20) is shown as a column, where different colors indicate that users preferred the corresponding method instead of the ground truth. As Scenes 21-30 do not provide the ground truth, we show the ratio of choosing the method that produces the most likable results.}
  \label{fig:userstudy}
\end{figure}

\noindent \textbf{Qualitative comparison: } In Fig.~\ref{fig:ball_comparions}, we visualize illumination maps and the corresponding rendering from several state-of-the-art methods and our pipeline. Gardner \emph{et al.}~\cite{Gardner17} regress a limited field-of-view photo to HDR illumination without strong assumptions on scene geometry, material properties, or lighting. As they predict illumination for the whole scene, this method cannot model spatial variations. 
EMLight~\cite{zhan2020emlight} ignores the complex scene geometry and simplifies the lighting distribution of scenes with a Gaussian map. Thus it can not deal with occlusion. In contrast, we estimate depth to have a better understanding of the scene, which leads to more accurate illumination estimation. Similar to our method, Neural Illumination~\cite{song2019neural} also decomposes the task into sub-tasks. However, it struggles to infer the position of lighting with a limited receptive field, especially when the input is terribly sparse. As Neural Illumination~\cite{song2019neural} has not shared code and model weights, we implemented and trained their network with our datasets. We can see that with a global understanding in local-to-global inpainting module, our method produces environment maps with accurate lighting and perceptually plausible details, ensuring realistic shading.

We further conduct a user study on 20 scenes of Garon \emph{et al.}~\cite{garon2019fast} and additional 10 scenes from the Internet.  As shown in Fig.~\ref{fig:inserted}, multiple virtual objects are inserted to these scenes.
For scenes of Garon \emph{et al.}~\cite{garon2019fast}, we relight bunny models with diffuse materials using the ground truth light probe and estimated illumination from Neural Illumination~\cite{song2019neural}, Li \emph{et al.}~\cite{li2020inverse} and our method. Li \emph{et al.}~\cite{li2020inverse} achieve the state-of-the-art performance by leveraging a deep inverse rendering framework to obtain a complete scene reconstruction, estimating shape, spatially-varying lighting, and non-Lambertian surface reflectance from a single RGB image. To reflect details of predictions, we render mirror spheres with estimated illumination in the remaining scenes.
The user study is conducted by asking 84 users to choose which rendering is more realistic between rendered image pairs. The results are shown in Fig.~\ref{fig:userstudy}. For scenes with inserted bunny models, Li \emph{et al}.~\cite{li2020inverse} and our method both beat each other in half of the scenes, which suggests that they are comparable in predicting lighting distribution. However, Li \emph{et al.}~\cite{li2020inverse} model the lighting with spherical Gaussian, leading the mirror sphere looks diffuse. For Scenes 21-30, our method outperforms the other methods on the mirror spheres, indicating that our method produces plausible details in coherence with the environment.
 
\begin{table}[tb]
\begin{center}
  \caption{Comparing the quantitative performance of our method to Gardner \emph{et al.}~\cite{Gardner17}, Neural Illumination~\cite{song2019neural} and EMLight~\cite{zhan2020emlight}. D, S, M denote a diffuse, a matte silver and a mirror material of the rendered spheres, respectively. A represents the angular error~\cite{apple-hdr}.}
  \label{tab:light_result}
\begin{tabular}{c|ccccc}
\multicolumn{2}{c}{\textbf{Metrics}}&~\cite{Gardner17}&~\cite{song2019neural}&~\cite{zhan2020emlight}&Ours\\ \hline
\multirow{2}{*}{D} & \textbf{MSE$\downarrow$} &35.77  & 22.67& 12.08 & \textbf{9.18}\\ 
 & \textbf{RMSE$\downarrow$} &53.22  &39.28 & 18.26 & \textbf{14.36}\\ \hline
\multirow{2}{*}{S} & \textbf{MSE$\downarrow$} &33.84  &16.45 & 17.05 &\textbf{9.75} \\ 
 & \textbf{RMSE$\downarrow$} &53.38  &24.09 & 31.79 & \textbf{22.32}\\ \hline
\multirow{2}{*}{M} & \textbf{MSE$\downarrow$} &34.28  &23.69 & 16.71 &\textbf{9.64} \\ 
 & \textbf{RMSE$\downarrow$} & 52.28 &41.44 & 30.86 & \textbf{21.89}\\ \hline
\multirow{2}{*}{A} & \textbf{mean$\downarrow$} & 1.31 &1.92 & 1.21 &\textbf{1.13} \\ 
 & \textbf{std$\downarrow$} & 0.34 &\textbf{0.16} & 0.34 & 0.34\\ 
\end{tabular}
\end{center}
\end{table}
\noindent \textbf{Quantitative comparison: }
To evaluate our performance on estimated illumination, we render three spheres with different materials: gray diffuse, matte silver and mirror with predicted illumination maps and ground truth. Then we use some standard metrics including root-mean-square error (RMSE) and the mean absolute error (MAE) to evaluate. To evaluate the accuracy of light sources, we take the mean angular error from both ground truth lights and predicted lights as the final angular error~\cite{apple-hdr} between the two HDR illumination map.
All these metrics have been widely adopted in the evaluation of illumination prediction. For evaluation, we use 2,000 pairs of input LDR images from Laval dataset and the ground truth HDR illumination maps captured at the camera. 
Tab.~\ref{tab:light_result} shows quantitative comparisons of our approach against Gardner \emph{et al.}~\cite{Gardner17}, Neural Illumination~\cite{song2019neural} and EMLight~\cite{zhan2020emlight}. 
As seen, our pipeline outperforms all methods in comparison under different evaluation metrics and materials. 

\begin{figure}[tb]
  \begin{center}
  \renewcommand\tabcolsep{1.0pt}
  \begin{tabular}{cccc}
  \small{Input image} & \small{~\cite{song2019neural}} & \small{Ours} & \small{Ground truth}\\
    \includegraphics[width=0.24\linewidth,   clip]{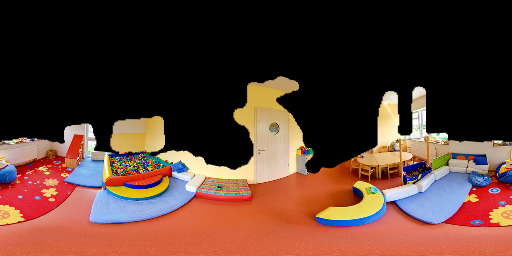}&
    \includegraphics[width=0.24\linewidth,   clip]{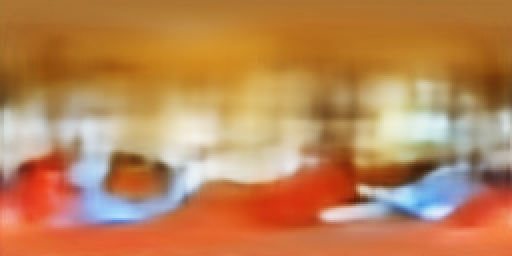}  &
    \includegraphics[width=0.24\linewidth,   clip]{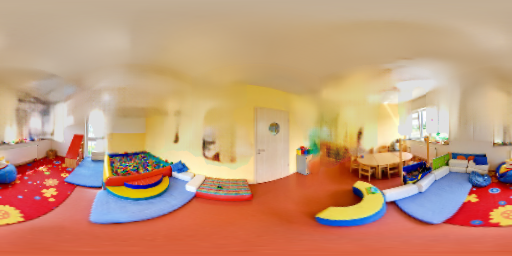}   &
    \includegraphics[width=0.24\linewidth,   clip]{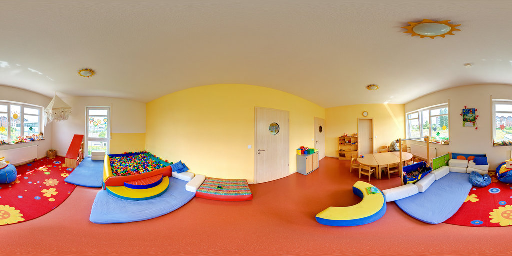} \\
    
    \includegraphics[width=0.24\linewidth,   clip]{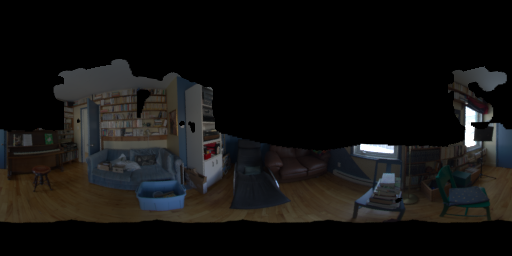}&
    \includegraphics[width=0.24\linewidth,   clip]{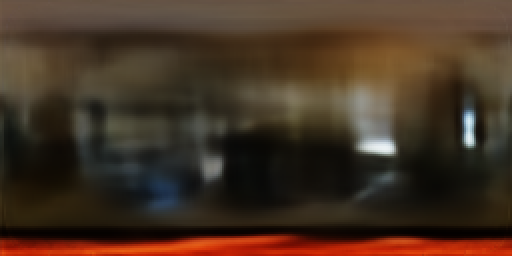}  &
    \includegraphics[width=0.24\linewidth,   clip]{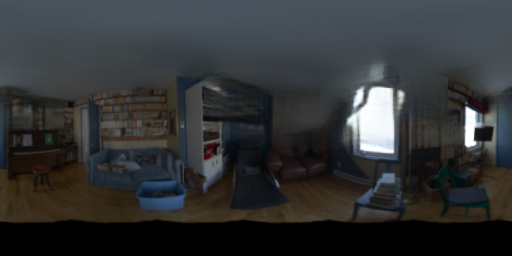}   &
    \includegraphics[width=0.24\linewidth,   clip]{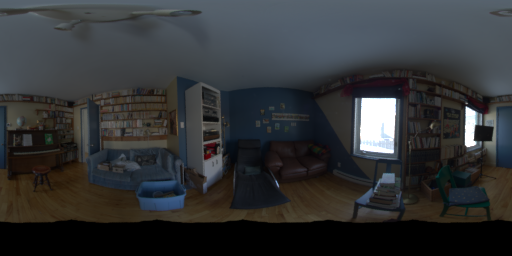}\\
  \end{tabular}
  \end{center}
  \caption{Qualitative comparison with Neural Illumination~\cite{song2019neural} on large-scale panorama inpainting. }
  \label{fig:envi_comparions}
\end{figure}
\subsection{Evaluation for panorama inpainting}

\begin{table}[t]
\begin{center}
  \caption{Quantitative comparisons on the panorama inpainting with Neural Illumination~\cite{song2019neural} and different model variants in the ablation study.}
  \label{tab:inpaint_result}
  \begin{tabular}{cccc}
    \textbf{Method}&  \textbf{SSIM$\uparrow$} & \textbf{PSNR$\uparrow$} & \textbf{FID$\downarrow$}\\
    \midrule
    Neural Illumination~\cite{song2019neural}  &0.30 &14.85 &255.60\\
    Ours  &\textbf{0.68}  &21.44 &\textbf{39.36}\\
    \midrule
    -Cubemap &0.62  &20.51 &70.09\\
    -Local  &0.60  &19.46 &112.32\\
    -GAN  &\textbf{0.68}  &\textbf{21.50} &76.07\\
\end{tabular}
\end{center}
\end{table}

\noindent \textbf{Qualitative comparison: }
To demonstrate the effectiveness of our transformer-based network for global inpainting, we compare inpainted results from Neural Illumination~\cite{song2019neural} and our method in Fig.~\ref{fig:envi_comparions}. Specifically, Neural Illumination ~\cite{song2019neural} could generally produce rough structures. However, the limited receptive fields of CNNs prohibit the understanding of global structures in the panorama. Besides, Neural Illumination doesn't maintain the visible regions with masks, hence these regions will be changed after prediction. In contrast, our results are more elegant with significantly fewer noticeable inconsistencies and artifacts, outperforming Neural Illumination~\cite{song2019neural} in panorama inpainting.

\noindent \textbf{Quantitative comparison: }
The inpainting evaluation is conducted on our test set containing 5,000 pairs of masked input and the ground truth. Tab.~\ref{tab:inpaint_result} shows quantitative comparisons of our approach against Neural Illumination~\cite{song2019neural}. Average PSNR, SSIM, FID values are listed for the inpainted LDR panoramas. 
Obviously, our method achieves superior results compared with Neural Illumination in all metrics. This further shows that our method achieves state-of-the-art performance on large-scale panorama inpainting.

\begin{figure}[t]
  \begin{center}
  \renewcommand\tabcolsep{1.0pt}
  \begin{tabular}{ccccccc}
  \multicolumn{4}{c}{\includegraphics[width=0.52\linewidth,   clip]{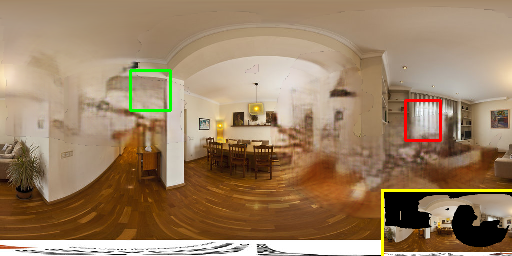}}&\includegraphics[width=0.13\linewidth,   clip]{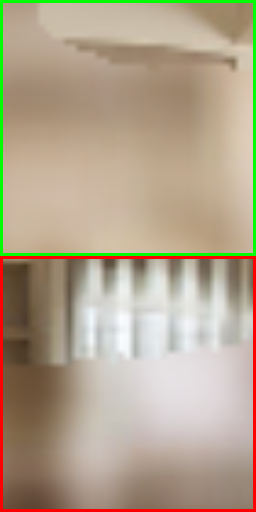}&
  \includegraphics[width=0.13\linewidth,   clip]{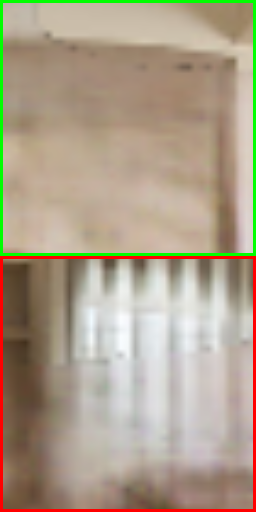}&
  \includegraphics[width=0.13\linewidth,   clip]{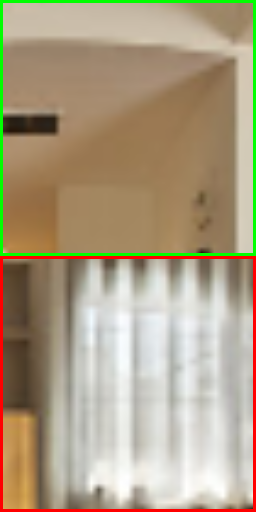}\\
  \multicolumn{4}{c}{\small{Our result and the input}}&\small{-GAN}&
  \small{Ours}&
  \small{Reference}\\
     \multicolumn{4}{c}{\includegraphics[width=0.52\linewidth,   clip]{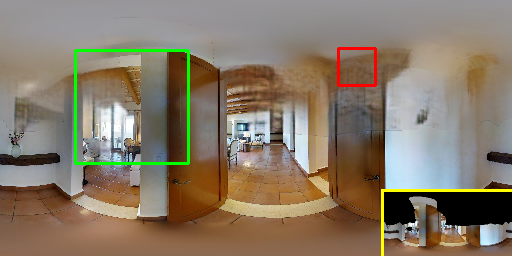}}&\includegraphics[width=0.13\linewidth,   clip]{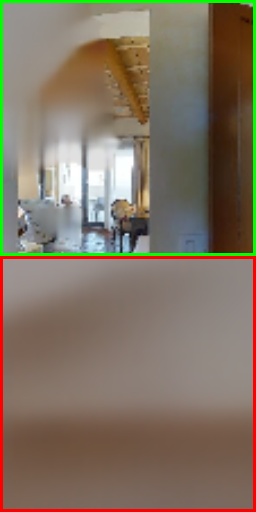}&
  \includegraphics[width=0.13\linewidth,   clip]{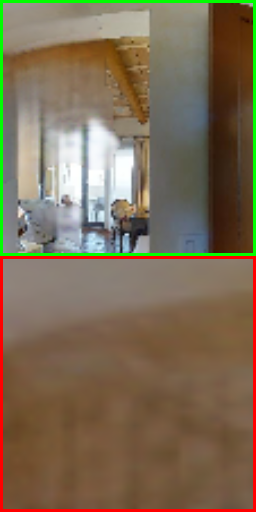}&
  \includegraphics[width=0.13\linewidth,   clip]{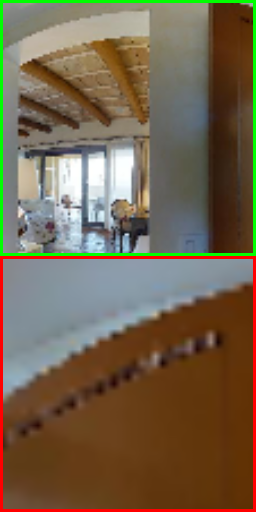}\\
  \multicolumn{4}{c}{\small{Our result and the input}}&\small{-Cubemap}&
  \small{Ours}&
  \small{Reference}\\
  \multicolumn{4}{c}{\includegraphics[width=0.52\linewidth,   clip]{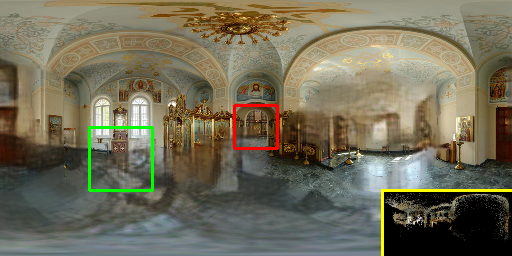}}&\includegraphics[width=0.13\linewidth,   clip]{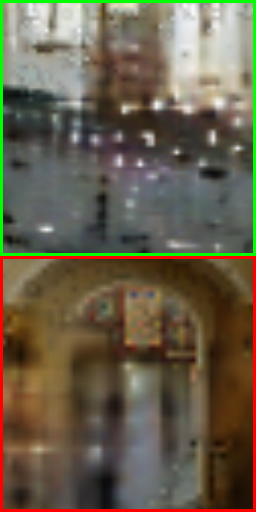}&
  \includegraphics[width=0.13\linewidth,   clip]{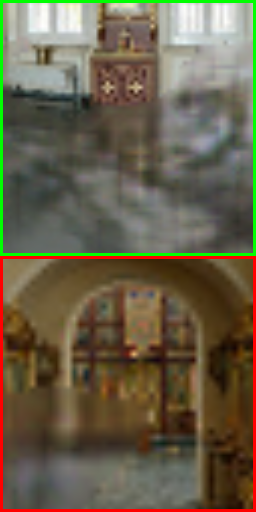}&
  \includegraphics[width=0.13\linewidth,   clip]{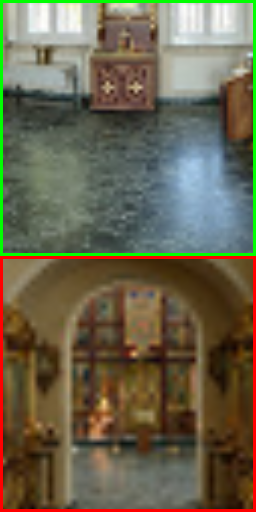}\\
  \multicolumn{4}{c}{\small{Our result and the input}}&\small{-Local}&
  \small{Ours}&
  \small{Reference}\\
  
  \end{tabular}
  \end{center}
  \caption{Validating the contribution of the GAN loss, the cubemap projection and the local inpainting module.}
  \label{fig:abstudy}
\end{figure}

\subsection{Ablation study}
\label{section:ablation}
To evaluate the effectiveness of our design in our method, we develop three model variants, denoted as -GAN model, -Cubemap model and -Local model respectively. The quantitative results are reported in Tab.~\ref{tab:inpaint_result}. We also qualitatively evaluate the performance of the model variants in Fig.~\ref{fig:abstudy}. 

-GAN represents our PanoTransformer trained without the GAN loss. From the top row in the Fig.~\ref{fig:abstudy}, we can see that our model trained without the GAN loss produces over-smooth textures that are close to the mean intensity of the surround regions. With the help of the GAN loss, our complete model is able to produce high-frequency signals and hallucinate realistic details.

Taking cubemap projection as the input aims to remove distortion in the panorama. To show the effectiveness of cubemap projection, we adapt PanoTransformer to -Cubemap model that takes the equirectangular projection as the input and outputs the LDR panoramas directly. As seen in the middle row in Fig.~\ref{fig:abstudy}, the -Cubemap model suffers from distorted structures. Our complete model outperforms the -Cubemap model clearly, demonstrating the superiority of the cubemap projections in handling spherical signals. 

To validate the importance of our local inpainting module, we remove this module and directly train PanoTransformer with sparse panoramas $\mathbf{\hat{P}}$. The bottom row in Fig.~\ref{fig:abstudy} shows that the -Local model introduces artifacts to the prediction, which explains the attention maps of the sparse input in Fig.~\ref{fig:attention}. With the local inpainting module, our pipeline produces more faithful and clearer results, indicating that the local inpainting module promotes the performance of PanoTransformer significantly.

\subsection{Limitation and discussion}
Since we decompose the illumination estimation into three subtasks, the quality of predictions is affected by the performance of these subtasks. Thanks to the tremendous success of deep learning, depth estimation and HDR panorama reconstruction from one single image have achieved significant progresses. We resort to the off-the-shelf DPT~\cite{dpt2021} for depth estimation and the pretrained network of Santos \emph{et al.}~\cite{LDR2HDR} for HDR panorama reconstruction. More details about the pretrained models can be found in the supplemental materials. Although we do not finetune for our tasks, the predicted depth and HDR map are still acceptable due to the robustness of the state-of-the-art networks. Nonetheless, we still have failure cases when the estimated depth is unreliable. Our performance can be improved with the advance of depth estimation and HDR panorama reconstruction.

\section{Conclusion}
Locale-aware indoor illumination estimation can be decomposed into three sub-tasks: depth-based image warping, panorama inpainting and HDR panorama reconstruction. In this paper, we propose a novel local-to-global panorama inpainting pipeline that well solves the second sub-task, significantly improving the performance of the spatially-varying indoor lighting estimation. In the pipeline, a depth-guided local inpainting module is first applied to fill holes due to pixel stretching. For global inpainting, we resort to a transformer-based network that captures distortion-free features from cubemap projection and enables physically-plausible reconstruction of global structures in the large out-of-view regions. We have validated our state-of-the-art performance through qualitative and quantitative comparisons against previous methods. Ablation studies further exemplify that the specially-designed large-scale panorama inpainting pipeline is able to produce perceptually plausible details coherent with the scene, leading to realistic shading after virtual object insertion.

{\small
\bibliographystyle{ieee_fullname}
\bibliography{acmart}
}

\end{document}